\documentclass[conference]{IEEEtran}
\usepackage{times}

\usepackage[numbers]{natbib}
\usepackage{multicol}
\usepackage[bookmarks=true]{hyperref}

\usepackage{amsmath,amssymb,amsfonts}
\usepackage{algorithmic}
\usepackage{graphicx}
\usepackage{textcomp}
\usepackage{xcolor}
\def\BibTeX{{\rm B\kern-.05em{\sc i\kern-.025em b}\kern-.08em
    T\kern-.1667em\lower.7ex\hbox{E}\kern-.125emX}}

\usepackage{graphicx}
\usepackage{caption}
\usepackage{subcaption}
\usepackage{url}
\newcommand{\emily}[1]{#1}
\newcommand{\andrew}[1]{#1}
\newcommand{\andrewhidden}[1]{}
\newcommand{\andrewrss}[1]{\andrewhidden{#1}}
\newcommand{\revision}[1]{#1}
\newcommand{\revisiontwo}[1]{#1}

\newcommand{\rsschange}[1]{\textcolor{black}{#1}}
\newcommand{\rsschangetwo}[1]{\textcolor{black}{#1}}
\newcommand{\rsschangethree}[1]{\textcolor{black}{#1}}

\newcommand{\name}[0]{MAVERIC}

\makeatletter
\newcommand{\linebreakand}{%
  \end{@IEEEauthorhalign}
  \hfill\mbox{}\par
  \mbox{}\hfill\begin{@IEEEauthorhalign}
}

\begin{document}

\title{MAVERIC: 
A Data-Driven Approach to Personalized Autonomous Driving }


\author{\IEEEauthorblockN{Mariah Schrum}
\IEEEauthorblockA{\textit{Institute for Robotics and Intelligent Machines} \\
\textit{Georgia Institute of Technology}\\
Atlanta, United States \\
mschrum3@gatech.edu}
\and
\IEEEauthorblockN{Emily Sumner}
\IEEEauthorblockA{\textit{Toyota Research Institute} \\
Los Altos, United States \\
emily.sumner@tri.global
}
\and
\IEEEauthorblockN{Matthew Gombolay}
\IEEEauthorblockA{\textit{Interactive Computing} \\
\textit{Georgia Institute of Technology}\\
Atlanta, United States \\
matthew.gombolay@cc.gatech.edu}
\linebreakand 
\IEEEauthorblockN{Andrew Best}
\IEEEauthorblockA{\textit{Toyota Research Institute} \\
Los Altos, United States \\
andrew.best@tri.global}
}


%

\maketitle

\begin{abstract}

Personalization of autonomous vehicles (AV) 
may significantly increase
trust, use, and acceptance. In particular, we hypothesize that the similarity of an AV's driving style compared to the end-user's driving style will have a major impact on  end-user's willingness to use the AV. To investigate the impact of driving style on user acceptance, we 1) develop a data-driven approach to personalize driving style  and \revision{2) demonstrate that personalization 
\andrew{significantly impacts} attitudes towards AVs.}  Our approach learns \revision{a high-level model that tunes low-level controllers to ensure safe and personalized control of the AV.} The key to our approach is learning an informative, personalized embedding that represents a user's driving style.
Our framework is capable of calibrating the level of aggression so as to optimize driving style based upon driver preference. Across two human subject studies (n = 54), we first demonstrate  our approach mimics the driving styles of end-users and can tune attributes of style (e.g., aggressiveness).  Second, we investigate the factors (e.g., trust, personality etc.) that impact homophily, i.e. an individual's preference for a driving style similar to their own. 
We find that our approach generates driving styles consistent with end-user styles ($p<.001$) and participants rate our approach as more similar to their level of aggressiveness ($p=.002$). We find that personality ($p<.001$), perceived similarity ($p<.001$), and high-velocity driving style ($p=.0031$) significantly modulate the effect of homophily.\looseness=-1

\end{abstract}

\IEEEpeerreviewmaketitle

\section{Introduction}

Driving style is defined as the characteristics of driving related to the judgment and decisions of the driver in a specific situation \cite{EBOLI2017945}. Research has shown that driving styles differ greatly amongst individuals \cite{differentStyles}. 
For example, the way in which a driver interacts with other drivers, the level of aggression that a driver exhibits, and tendency to commit traffic violations are characteristics that define an individual's unique driving style.
Because of these individual differences, when riding in an autonomous vehicle (AV), end-users' expectations and preferences for the behavior of the AV will likely be influenced by their own driving style \cite{hasenjager,hoedemaeker,Sun2020}. One-size-fits-all models employed by AVs which ignore driver differences may lead to decreased acceptance  \cite{hasenjager}. Instead, the driving style of AVs should be personalized to fit the preferences and expectations of individual end-users.

Much of the prior work in optimizing AV driving styles has assumed that, to increase end-user acceptance and trust, AVs should mimic  users' unique driving styles \cite{Sun2020,Ekman2019ExploringAV}.  However, even if we are able to personalize an AV's behavior, not all end-users will necessarily want the AV to drive \textit{exactly} as \revision{the end-user drives} \cite{Yusof,Basu}. 
In fact, prior work has suggested that some end-users may want an AV to drive more cautiously than they drive \cite{Ekman2019ExploringAV,Yusof,Basu}. Additionally, factors such as trust and familiarity with AVs and various personality traits may affect  preference for driving styles similar to one's own \cite{EBNALI2021103226,Choi,PersonalityandStyle}. 

\begin{figure}[t!]
    \centering
    \includegraphics[width=0.48  \textwidth]{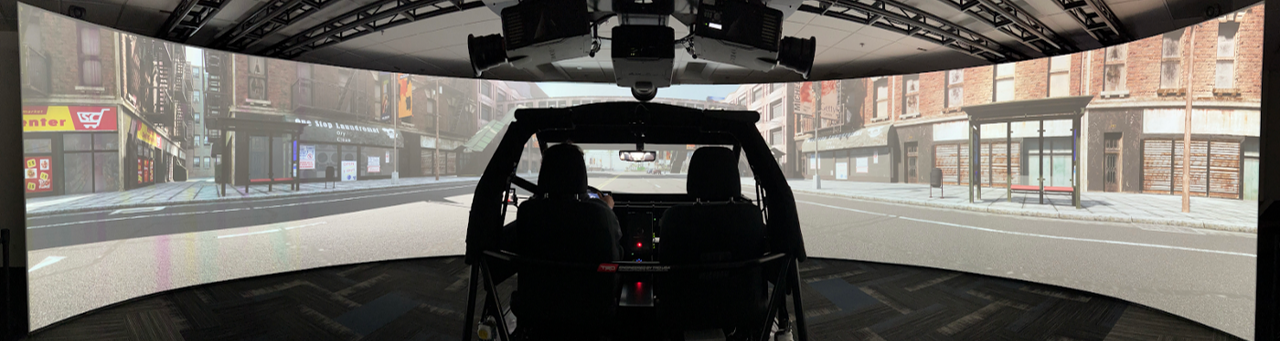}
    \caption{ 6-DOF driving simulator 
    }
    \label{fig:sim}
    \vspace{-.3cm}
\end{figure}

 Based upon evidence from prior work \cite{EBNALI2021103226,Choi,PersonalityandStyle,Ekman2019ExploringAV}, we hypothesize that an individual's optimal driving style is a function of both the end-user's own driving style and various subjective factors, such as personality. Therefore, to optimize driving style, an AV must be capable of learning about an end-user's own driving style as well as modulating this style based upon relevant end-user characteristics. \rsschange{We identify two key shortcomings in prior work. First, prior approaches that are capable of modulating the driving style of an AV  do not take into account the end-user's own driving style \cite{Eriksson2015,Bae2019}, despite prior work indicating that end-user's own driving style is an important predictor of the optimal driving style. Second, prior works which  mimic an end-user's driving style \cite{Bolduc, Kuderer} are not capable of modulating various aspects of driving style to account for end-users who do not wish the driving style to exactly mirror their own.}  
 
 In this work, we develop a data-driven approach called Manipulating Autonomous Vehicle Embedding Region for Individuals' Comfort (\name{}) which is capable of producing 
 \andrew{an optimized} driving style for an end-user. This driving style  is based upon the end-user's own driving style but can also be modulated to adjust the level of aggression. 
 \revision{By observing the driving of an end-user, \name{}  learns a  high-level model via a neural network architecture that predicts personalized control parameters for low-level controllers to \rsschange{mimic the driving style of the end-user.}} Simultaneously, \name{} learns a personalized embedding representing the driving style of an end-user. By shifting the personalized embedding along the gradient of aggression, \revision{\name{} is capable of tuning the AV driving style} to be more aggressive or cautious \rsschangetwo{while maintaining other  characteristics such as minimum headway distance and fraction of time in left lane.} This capability allows us to modulate the AV's level of aggressiveness with respect to an end-user's style so as to optimize the AV's driving style. \looseness=-1

In a  human-subjects study, we investigate if \name{} can effectively mimic an individual's driving style as well as modulate aggression \rsschange{\emph{with respect to one's own driving style} while maintaining other aspects of driving style}.  Additionally, we investigate the factors that influence the effect of \textit{homophily}, i.e., preference for a driving style similar to one's own. We demonstrate that preferred driving style is related both to one's own style as well as  \revision{personality traits, perceived similarity, and high-velocity driving}.

In this work we contribute the following:

\begin{enumerate}
    \item We formulate \name{}, a novel framework to personalize driving style and modulate aggressiveness while maintaining other aspects of driving style. 
    \item We demonstrate  that \name{} can closely match an end-user's driving style ($p<.001$) as well as produce more aggressive ($p<.001$) and more cautious ($p<.001$) driving in a high-fidelity driving simulator. 
    \item We find that personality
($p<.001$), perceived similarity ($p<.001$), and high-velocity driving style ($p=.0031$)  significantly impact the effect of homophily.
\end{enumerate}

\section{Related Work}

\revision{Researchers have demonstrated that personalization of AVs can lead to increased acceptance \cite{Ekman2019ExploringAV,Sun2020} and   may decrease motion sickness \cite{ISKANDER2019716}.
To achieve this objective of personalization,}
Kuderer et al. utilized an inverse reinforcement learning approach to produce personalized AV behavior  via a learned cost function \cite{Kuderer}. Ling et al. adapted driving style online based on the emotional responses of passengers \cite{Emotion}. Other work investigated personalization of specific aspects of driving \cite{Bolduc,Feng2022}.  For example, Bolduc et al.  developed an approach to match driver's style for adaptive cruise control \cite{Bolduc}.\looseness=-1  

Yet, prior work suggests that end-users may not  want an AV to drive \emph{exactly} as they drive and various factors may influence a driver's preference for a specific driving style \cite{Basu,Ekman2019ExploringAV,Yusof}. \rsschange{Prior work has also shown that specifically the level of aggression of an AV has a large impact on preference  \cite{Basu,Ekman2019ExploringAV,Yusof}.} For example, Ekman et al. found that a defensive driving style produced higher trust scores in a Wizard-of-Oz  study \cite{Ekman2019ExploringAV}.
Yusof et al. also found defensive driving  was preferred, even more so by aggressive drivers \cite{Yusof}.
Basu et al. also found that participants \emph{did not} want AVs to drive as they drive, instead preferring the AV to drive like the end-user \textit{thinks} they drive, which often differs from their actual driving style \cite{Basu}.
The authors in \cite{Basu} suggest that we can not simply rely on mimicry  and instead, must also account for other end-user characteristics to determine an optimal driving style. \looseness=-1

\andrewrss{I have added a benefit here. Please review and discard if you disagree, I am trying to reflect that our users don't need to know how to tune a speed profiler as is done in Bae}
\rsschange{Prior work has introduced several classic control approaches which allow for tuning of the level of AV aggression. For example, Eriksson and Svensson \cite{Eriksson2015} introduced a linear quadratic controller for tuning the driving of an AV to optimize ride quality. Bae et al. \cite{Bae2019} introduced an approach which allows the user to specify the desired parameters for a controller to adjust the driving style of the vehicle with respect to acceleration and jerk. However, these approaches do not consider tuning driving style with respect to the end-user's own style and may require expert knowledge of vehicle dynamics and complex control parameters to determine the correct parameter settings.}

While prior work has investigated mimicking driving style and tuning style via controllers, no prior work has created an architecture that can modulate driving style \textit{with respect to} an end-user's own style. Yet, prior work provides evidence that this functionality is important for optimizing driving style for an end-user \cite{Basu,Yusof}. Additionally, prior work has not extensively investigated the relationship between subjective factors and preference for styles similar to one's own. In our work, we seek to fill these gaps by proposing an approach capable of producing more or less aggressive behavior with respect to the end-user's own driving style. Additionally, we conduct a thorough investigation into the factors that impact the effect of homophily.

\begin{figure}[t]
    \centering
    \includegraphics[width=0.48  \textwidth]{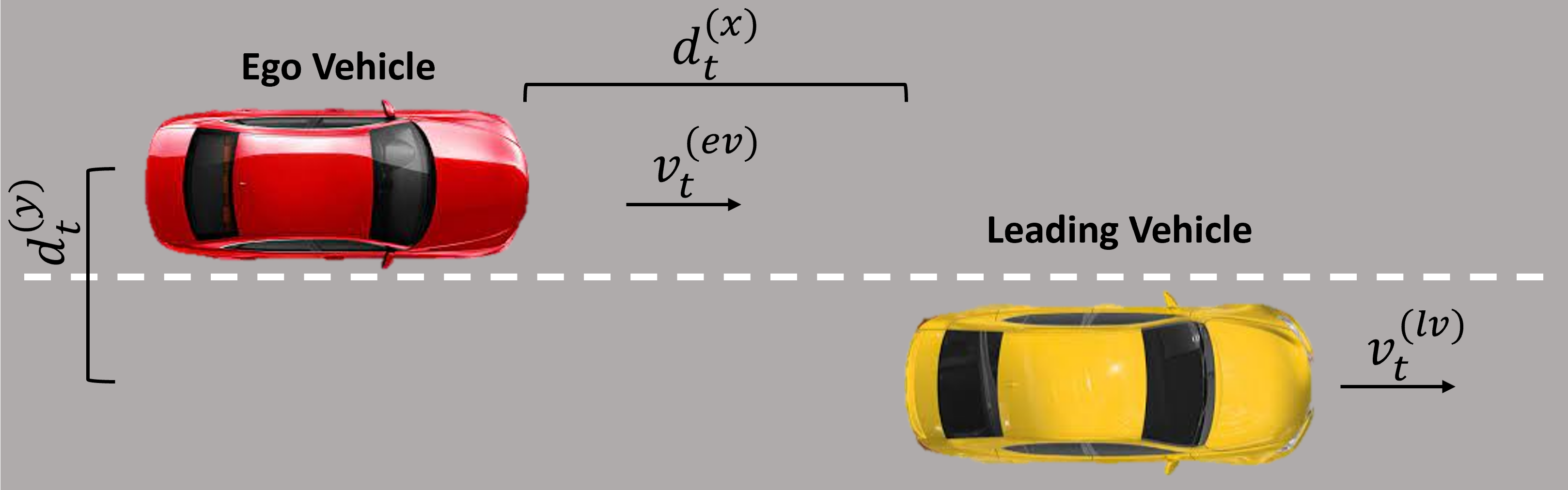}
    \caption{This figure shows our domain of light traffic and associated state information. $v^{(ev)}_t$ is the velocity of the ego  at time $t$, $v^{(lv)}_t$ the velocity of the leading vehicle, $d^{(x)}_t$ the distance between the leading vehicle and ego  in the $x$ direction at time $t$, and $d^{(y)}_t$ the distance  in $y$ at time $t$. 
    }
    \label{fig:domain}
\end{figure}

\section{Methodology}
In the following section we  provide an overview of \name{}. We discuss our architecture 
and how we endow our framework with the ability to both mimic an end-user's driving style as well as modulate  aggression. Fig. \ref{fig:domain} depicts the state information relevant to our architecture.

\begin{figure*}[t]
    \centering
    \includegraphics[width=  \textwidth]{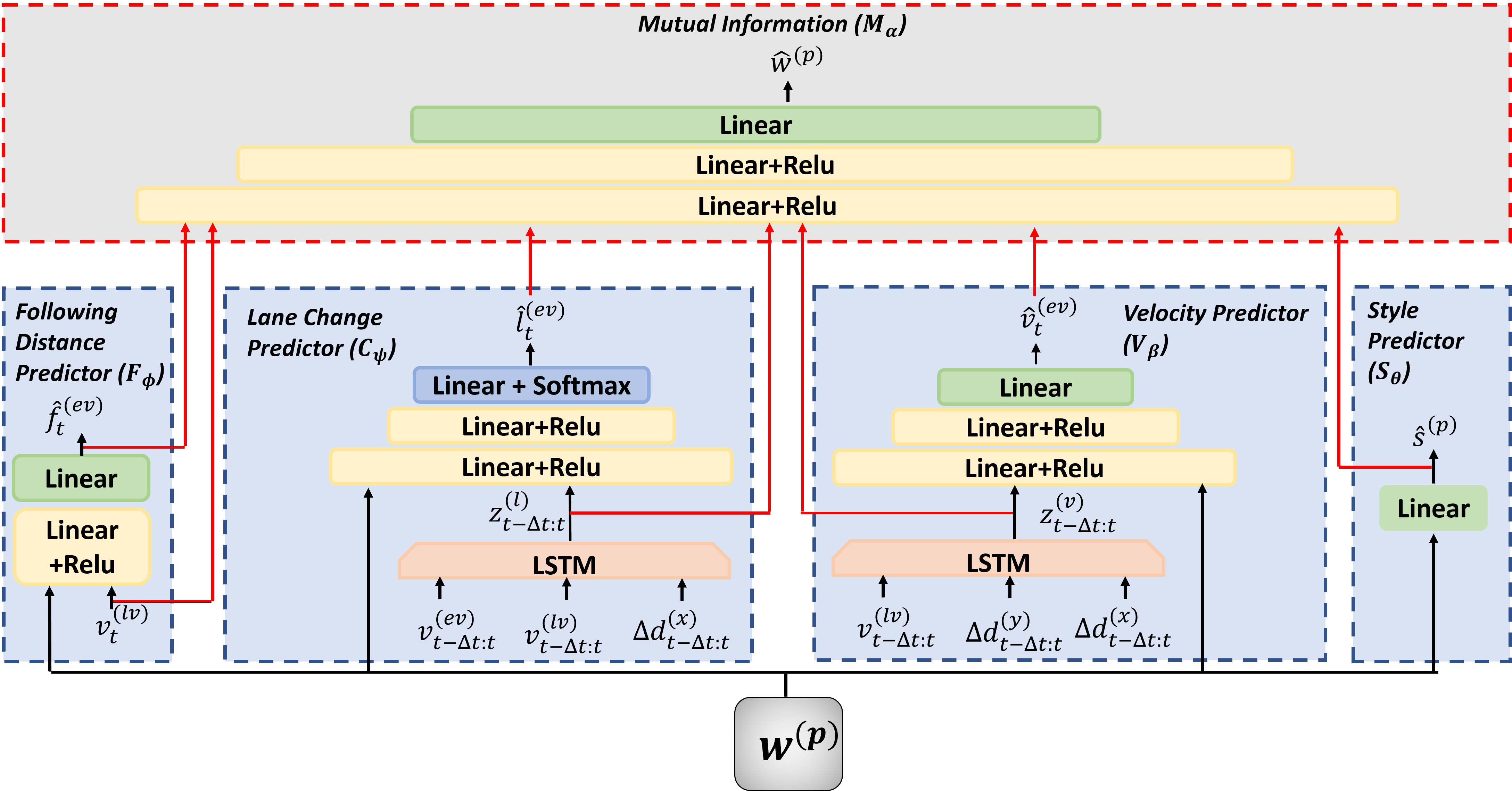}
    \caption{ This figure shows our network architecture. $F_\phi$ predicts the following distance. $C_\psi$ predicts when a lane change should occur for the ego vehicle. $V_\beta$ outputs the velocity of the ego vehicle.  $S_\theta$ is the style predictor subnetwork which predicts the subjective aggressive style of the participant from the personalized embedding, $w^{(p)}$. $v^{(ev)}_{t-\Delta t:t}$ is the ego velocity and  $v^{(lv)}_{t-\Delta t:t}$ is the velocity of the lead vehicle from time $t-\Delta t$ to $t$.  $d^{(x)}_{t-\Delta t:t}$ is the distance between the ego and leading vehicle in $x$ and $d^{(y)}_{t-\Delta t:t}$ is the distance in $y$. \rsschange{$\hat{w}^{(p)}$ is the estimate of the participant's personalized embedding sampled from the approximate posterior defined by $M_\alpha$.}}
    \label{fig:architecture}
\end{figure*}

\subsection{Network Architecture}
Our network architecture is depicted in Fig. \ref{fig:architecture}. Our network simultaneously learns the high-level parameters of low level controllers and the personalized embedding, $w^{(p)}$, representing the driving style of an individual, $p$. See Section \ref{sec:lowLevelControllers} for details on the low-level controllers. Our network is composed of five subnetworks: the Following Distance Predictor, $F_\phi$,  Lane Change Predictor, $C_\psi$, Velocity Predictor, $V_\beta$,  Style Predictor, $S_\theta$, and Mutual Information, $M_\alpha$. 

\noindent\textbf{Following Distance Predictor ($F_\phi$)} 
\vspace{-.1cm}
\begin{itemize}
    \item Inputs: personalized embedding,  $w^{(p)}$, and   the velocity of the lead vehicle, $v^{(lv)}_{t}$. 
    \item Outputs: desired following distance, $\hat{f}^{(ev)}_t$, between the ego vehicle and the lead vehicle.
    \item Layers: \rsschangethree{fully-connected} with ReLU activations.
    \item Loss Function: mean-squared error loss defined as $L_1(\phi,w^{(p)})=\frac{1}{N}\sum_t ||f^{(ev)}_t-\hat{f}^{(ev)}_t\big||_2^2$), where $f^{(ev)}_t$ is the  ground truth following distance of the end-user.
\end{itemize}

\noindent\textbf{Lane Change Predictor ($C_\psi$)} 
\vspace{-.1cm}
\begin{itemize}
    \item Inputs: personalized embedding, $w^{(p)}$,  ego velocity, $v^{(ev)}_{t-\Delta t:t}$, the velocity of the lead vehicle, $v^{(lv)}_{t-\Delta t:t}$, and  x-distance between the ego and lead vehicle, $d^{(x)}_{t-\Delta t:t}$, from time $t-\Delta t$ to $t$.
    \item Outputs: $\hat{l}_t^{(ev)}$, i.e. when a lane change should occur.
    \item Layers:  \rsschangethree{fully-connected} with ReLU  and softmax activation.
    \item Loss Function:  cross-entropy loss defined as $L_2(\psi,w^{(p)})=-\frac{1}{N}\sum_t l_t^{(ev)} \log \hat{l}_t^{(ev)}$ where $l^{(ev)}_t$ is a binary variable indicating a lane change.
\end{itemize}


\noindent\textbf{Velocity Predictor ($V_\beta$)} 
\vspace{-.1cm}
\begin{itemize}
    \item Inputs: personalized embedding, $w^{(p)}$, velocity of the lead vehicle, $v^{(lv)}_{t-\Delta t:t}$, y-distance, $d^{(y)}_{t-\Delta t:t}$, and x-distance, $d^{(x)}_{t-\Delta t:t}$, between the ego and lead vehicle, from time $t-\Delta t$ to $t$.
    \item Outputs: predicted speed of the ego, $\hat{v}_t^{(ev)}$, at time $t$.
    \item Layers:  \rsschangethree{fully-connected} with ReLU activations.
    \item Loss Function:  mean-squared error loss defined as $L_3(\beta,w^{(p)})=\frac{1}{N}\sum_t ||v^{(ev)}_t-\hat{v}^{(ev)}_t\big||_2^2$) where $v^{(ev)}_t$ is the velocity of the end-user.
\end{itemize}


\noindent\textbf{Style Predictor ($S_\theta$)} 
\vspace{-.1cm}
\begin{itemize}
    \item Inputs: personalized embedding, $w^{(p)}$.
    \item Outputs: subjective aggressiveness, $\hat{s}^{(p)}$, of participant, $p$.
    \item Layers: \rsschangethree{fully-connected}.
    \item Loss Function:  $L_4(\theta,w^{(p)})=\frac{1}{N}\sum_p ||s^{(p)}-\hat{s}^{(p)}\big||_2^2$) where $s^{(p)}$ is the subjective aggressive style of the end-user as self-reported via a questionnaire. In Section \ref{sec:modulating}, we discuss \revision{the importance of this} subnetwork  and how we obtain $s^{(p)}$.
\end{itemize}


\textbf{Mutual Information ($M_\alpha$): } $M_\alpha$ seeks to maximize mutual information between the driving style of the individual and $w^{(p)}$  to ensure that the learned embedding captures the differences in driving style between individuals \rsschange{and that unique driving styles are represented by unique embeddings \rsschangetwo{\cite{Paleja2019,MINDMELD}}.} 
\vspace{-.1cm}
\begin{itemize}
    \item Inputs: $v_t^{(lv)}$, encodings of the relevant time series state information $z^{(l)}_{t-\Delta t:t}$ and $z^{(v)}_{t-\Delta t:t}$, and the outputs of each of the other subnetworks, $\hat{l_t}^{(ev)}$, $\hat{v}^{(ev)}_t$, $\hat{f}^{(ev)}_t$, and $\hat{s}^{(p)}$.
    \item Outputs: $\hat{w}^{(p)}\sim \mathcal{N}(\mu^{(p)},\sigma^{(p)})$.
    \item Layers:  \rsschangethree{fully-connected layers} with ReLU activations.
    \item Loss Function:  mean-squared error loss defined as $L_5(\alpha,w^{(p)})=\frac{1}{N}\sum_p ||w^{(p)}-\hat{w}^{(p)}\big||_2^2$).
\end{itemize}
\color{black}

\andrewrss{Still see that Z^l only really appears here. Do we need to be more explicit about the encoding, or is this a term of art that our ICRA reviewers just didnt know? Should we call out Z in figure 3?}
\rsschange{With this setup, we train $w^{(p)}$ to capture salient information about an end-user's driving style by maximizing a lower bound, $L_I(F_\phi,C_\psi,V_\beta,S_\theta,M_{\alpha})$, on mutual information. Eq. \ref{eq:mutualinfo} shows the lower bound on mutual information  as derived in Chen et al. \cite{Chen2016}. $\mathcal{X}$ represents the vector containing the relevant state parameter ($v_t^{(lv)}$, $z^{(l)}_{t-\Delta t:t}$, and $z^{(v)}_{t-\Delta t:t}$). $\mathcal{P}$ represents the vector containing the outputs of the subnetworks ($\hat{f^{(ev)}_t}$, $\hat{l^{(ev)}_t}$, $\hat{v^{(ev)}_t}$, and $\hat{s^{(p)}}$). }

\footnotesize
\begin{gather}
    I({w}^{(p)};{\mathcal{X}},\mathcal{P}) = H({w}^{(p)})-H({w}^{(p)}|\mathcal{X},\mathcal{P})  \nonumber
    \\  \geq \mathop{\mathbb{E}}[log(M_\alpha({w}^{(p)}|\mathcal{X},\mathcal{P}))]+H({w}^{(p)})  = L_I(F_\phi,C_\psi,V_\beta,S_\theta,M_{\alpha}) \label{eq:mutualinfo}
\end{gather}
\normalsize

\subsection{Training}
\rsschange{We train our \name{} architecture to minimize the five loss functions, $L_1(\phi,w^{(p)})$, $L_2(\psi,w^{(p)})$, $L_3(\beta,w^{(p)})$, $L_4(\theta,w^{(p)})$, and $L_5(\alpha,w^{(p)})$. Loss functions $L_1$ through $L_4$ are utilized to train the four predictor subnetworks. $L_5(\alpha)$ minimizes the MSE between the embedding $\hat{w}^{(p)}$ sampled from the approximate posterior and the true embedding, $w^{(p)}$ ($L_5(\alpha,w^{(p)})=\frac{1}{N}\sum_p ||w^{(p)}-\hat{w}^{(p)}\big||_2^2$).  This is equivalent to maximizing the log likelihood of the posterior represented by $M_\alpha$ \cite{Paleja2019}. We note that we initialize $w^{(p)}$ by sampling from the prior,  $\hat{w}^{(p)} \sim \mathcal{N}(0,1)$ \cite{Paleja2019}.  The sum of these five  losses is then backpropagated through each of the subnetworks and $w^{(p)}$ to simultaneously learn the personalized control parameters and the personalized embedding representing driving style.}


\subsection{Modulating Aggression}
\label{sec:modulating}

We designed \name{} to be capable of both matching driving styles of individuals and modulating aggression with respect to an individual's driving style.
Because \name{} learns a latent embedding space, we aim to create a dimension of aggression within the embedding space, allowing us to shift an embedding along that dimension and modulate aggression, while keeping other driving characteristics  unrelated to aggression constant.
To achieve this objective, we add an additional signal when learning the embedding space.
We add a network head, $S_\theta$, composed of a  \rsschangethree{fully-connected layer} which takes as input the personalized embedding, $w^{(p)}$.
$S_\theta$ is trained to predict the subjective aggressive driving style of the user as measured by the participant's response to the Aggressive Driving Behavior (ADB) scale \cite{AggressiveScale}. 
By doing so, we create an aggressive dimension within the embedding space.
We can then move along the gradient of aggression \rsschangetwo{($\nabla_w S_\theta$)} to produce a more or less aggressive driving style  as shown in Fig. \ref{fig:embeddings}.\looseness=-1

While driving style has multiple dimensions \cite{TAUBMANBENARI2016179}, we focus on  the aggressive dimension, as prior work has shown that this dimension has a large impact on end-user preference \cite{Basu,Ekman2019ExploringAV,Yusof}.
Other characteristics of driving could be modulated  by following a similar procedure.
While we acknowledge that the ADB scale is a noisy metric,
 as discussed in Section \ref{sec:results}, our results demonstrate that our method can effectively produce more and less aggressive behavior. Additionally, we find  we can represent the driving style of an end-user via a 3-dimensional vector and that increasing the size of the embedding did not significantly improve the accuracy of predictions.\looseness=-1

\subsection{Low Level Controllers}
\label{sec:lowLevelControllers}

MAVERIC learns the parameters for low-level controllers (e.g., velocity, timing of lane change, etc.) rather than directly learning the low-level control inputs (i.e., throttle and steering) to enable safety constraints and account for unexpected or dangerous behavior that could be produced by the network. For example, by learning the desired following distance for an end-user and utilizing an adaptive cruise controller to maintain this distance, we can ensure that the following distance remains safe. Additionally, by predicting when a lane change should occur via the neural network and utilizing a low-level  controller to execute the lane change, we ensure consistent and smooth lane changes. Furthermore, this hierarchical method of learning and control has been shown to produce better results in prior work \cite{Nasiriany2021}. 
\rsschangetwo{We utilized the specific controllers described below because they have been shown to be robust and produce desired behavior in prior work \cite{Snider2009}. However, these controllers could be exchanged for other controllers.}

\textbf{Lane Change Controller:} Our lane change controller is based on a Stanley controller \cite{Snider2009} and follows a Bezier curve \cite{Bae}. We compute the Bezier curve based upon the desired distance (selected to produce natural behavior)  to complete the lane change, while ensuring that the ego vehicle will not collide with the leading vehicle. \revision{The lane change controller executes a lane change when $\hat{l}^{(ev)}_t>\delta$.}

\textbf{Velocity Controller: } We utilize a proportional and integral (PI) controller to maintain the desired velocity, \revision{$\hat{v}_t^{(ev)}$}, of the ego vehicle as predicted by the neural network.

\textbf{Following Distance Controller:} When the distance between the ego and leading vehicle falls below threshold, $\lambda$, we switch from the velocity controller to the following distance controller. The following distance controller is a PI controller that  minimizes both the error between the desired following distance, \revision{$\hat{f}_t^{(ev)}$}, as predicted by the neural network and the difference in speed between the ego and leading vehicle subject to safety constraints on following distance.

\section{Human Subjects Studies}

We conducted two human subjects studies: A Model Training \revisiontwo{Study} (Study 1) and a Model Testing \revisiontwo{Study} (Study 2). In Study 1, we collect data from 30 participants to train \name{} and learn  $\theta$, $\phi$, $\psi$, $\alpha$, and $\beta$, and the participants' embeddings. We then freeze these parameters, and in Study 2, we collect driving data from 24 participants to learn their  embedding. Then each participant experiences the four AV conditions (Section \ref{sec:conditions}). Research was  approved by an IRB.


\subsection{Driving Simulator}
To test the abilities of \name{}, we utilize a high-fidelity, research grade driving simulator. 
 The simulator (Fig. \ref{fig:sim}) is an immersive 6-DOF platform capable of emulating the motion of a vehicle. The simulator is based on CARLA \cite{Dosovitskiy17}, ROS2, and Unreal Engine.\looseness=-1

\begin{figure}[t]
    \centering
    \includegraphics[width=0.45  \textwidth]{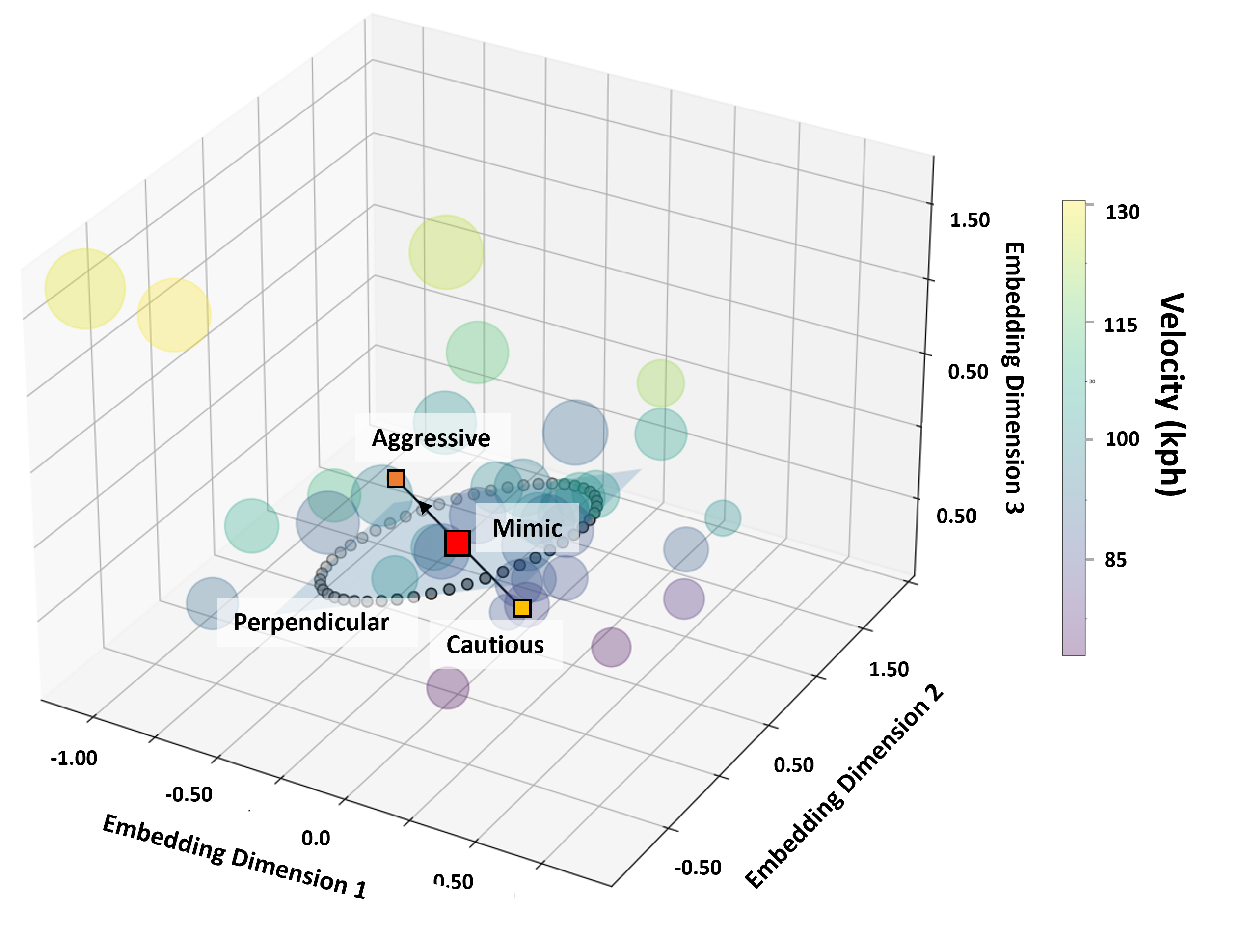}
    \caption{This figure shows the learned embedding space. The size of the points represents the subjective aggressive style of the participant and color represents the average velocity. The black line shows the vector of the aggressive gradient. The red square represents a candidate learned embedding of a participant. 
    We shift the embedding along the gradient to increase (orange square) or decrease (yellow square) the ADB score by 15 points to produce behavior for the aggressive and cautious conditions respectively.
   We randomly sample from the gray points to produce the Perpendicular behavior.\looseness=-1}
    \label{fig:embeddings}
\end{figure}

\subsection{Participants} 
\textbf{Model Training \revisiontwo{Study} (Study 1): } We recruited 30 participants (Mean age 35.4; 27\% Female)
 via word of mouth and mailing lists. Four of the participants were professional drivers who demonstrated aggressive, cautious, and their own driving style. In total, we collected 38 data points representing various driving styles. 

\textbf{Model Testing \revisiontwo{Study} (Study 2): } 
Study 2 was run with two different populations of participants to increase diversity. For the \emph{internal} study, 12 subjects were recruited internally 
(Mean age 34.42; 33.4\% Female). For the \emph{external} study, 12 subjects (Mean age 43.92; 41.7\% Female) were recruited from the general public via Fieldwork recruiting. External participants were compensated \$250. The populations are analyzed as a collapsed dataset because the procedure was identical.


\subsection{Procedure}
We investigate personalization of driving styles in the domain of light traffic  on a \rsschange{two-lane divided highway with both lanes going in the same directio}n (Fig. \ref{fig:domain}). 

\textbf{Study 1:}  Participants control the vehicle and demonstrate their driving style for 10 minutes. Their task is to drive as they would in their own vehicle. They are instructed to maintain the speed they would typically drive if the speed limit is 55mph and to pass other vehicles when they feel it is appropriate. In this domain, participants encounter vehicles in the same lane (lead vehicles) and in the adjacent lane (off-lane vehicles). Participants must make decisions about changing lanes, following distance, and velocity. The speed of the lead vehicles is randomly selected without replacement from the set \{$0.85v_e$, $0.9v_e$, $0.97v_e$, $0.9s$, $s$, $1.1s$\} where $v_e$ is the ego target speed and $s$ the posted speed (55mph). These speeds ensure consistency across participants but also ensures that a portion of the leading vehicles are slower than the ego, thus forcing the participant to make a decision about changing lanes. \looseness=-1

Participants first complete pre-study surveys to collect information about demographics and attitude towards AVs (Section \ref{sec:metrics}). Participants complete a practice session to familiarize themselves with the vehicle controls and domain.  We next collect driving data from the participants to learn the network parameters and their personalized embeddings. 

\begin{figure*}[t]
\centering
\begin{subfigure}{0.3\textwidth}
\includegraphics[width=\textwidth]{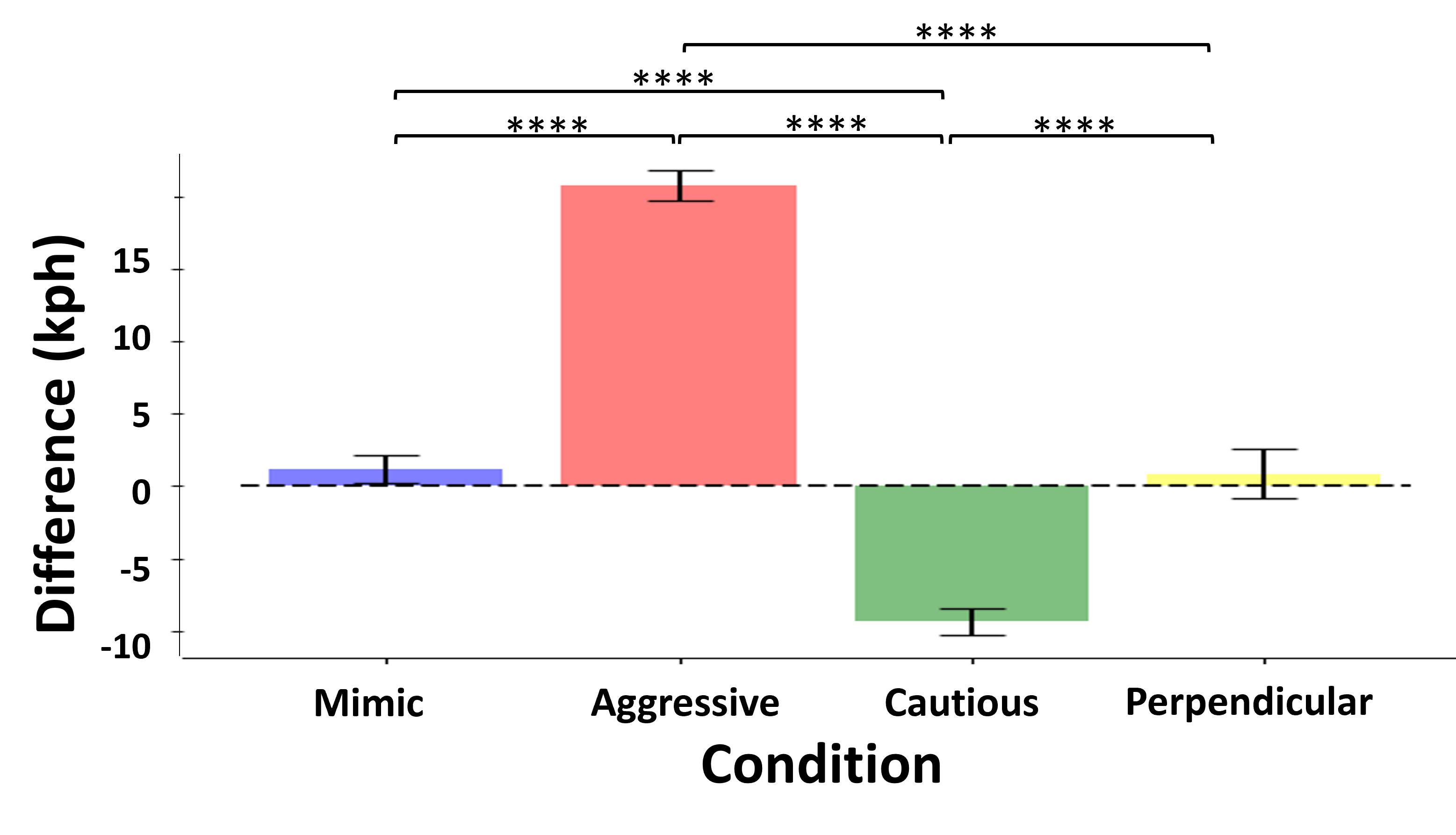}\vspace{-.2cm}
\caption{\footnotesize Mean velocity.} \label{fig:average_vel}
\end{subfigure}\hfil
\begin{subfigure}{0.3\textwidth}
\includegraphics[width=\textwidth]{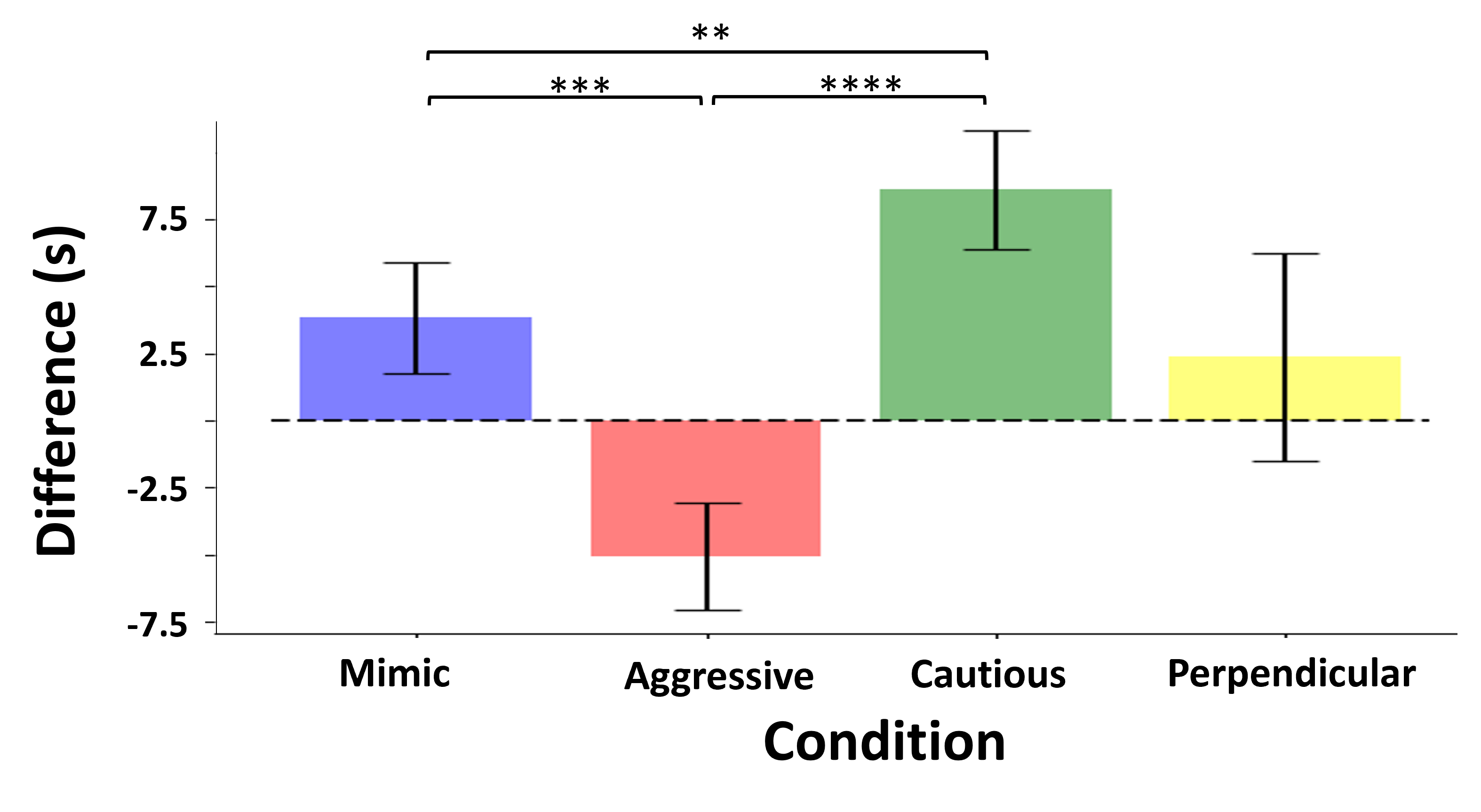}\vspace{-.2cm}
\caption{\footnotesize Mean headway time.} \label{fig:time_headway}
\end{subfigure}\hfil
\begin{subfigure}{0.3\textwidth}
\includegraphics[width=\textwidth]{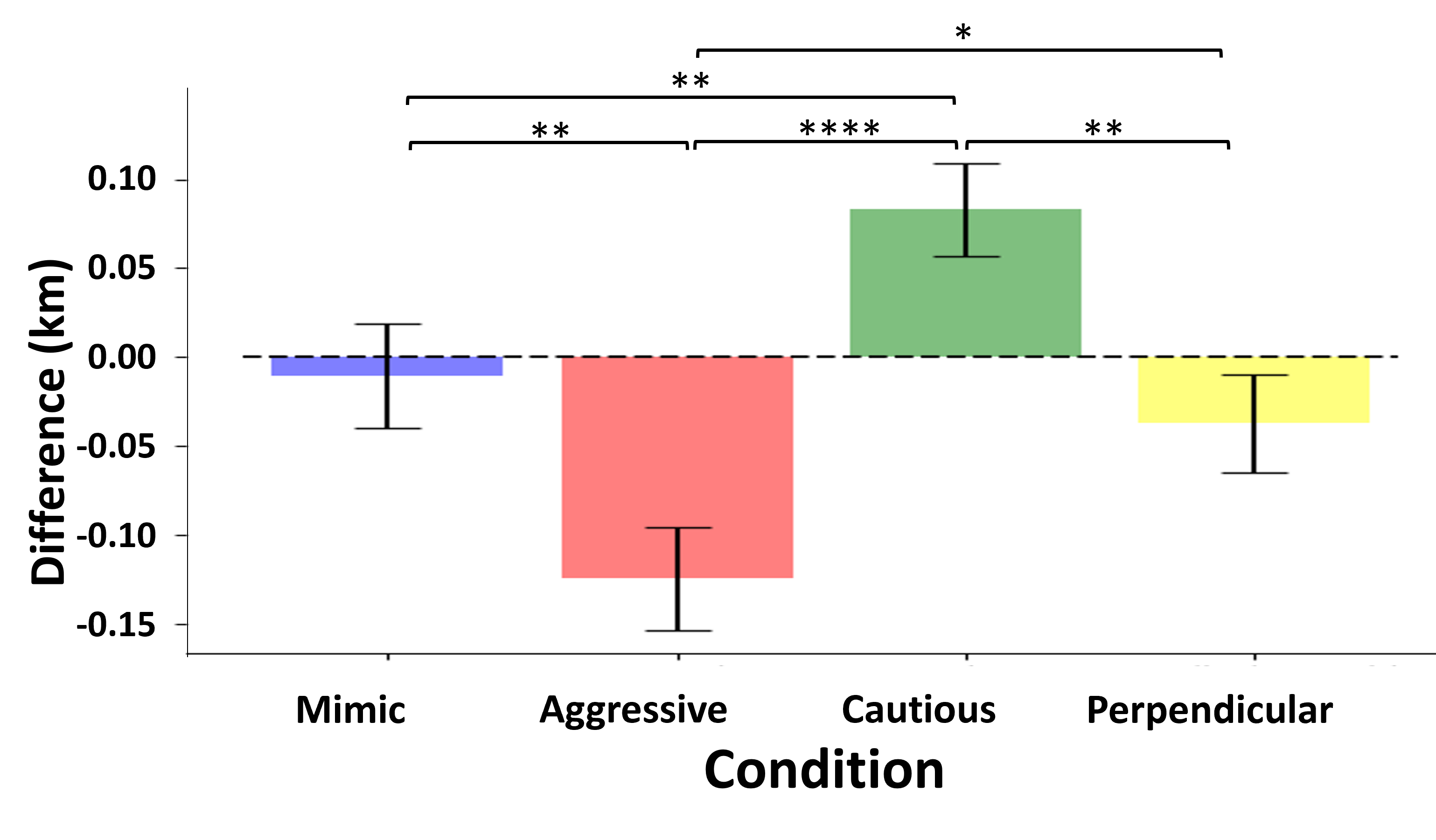}\vspace{-.2cm}
\caption{\footnotesize Distance headway merge back.}\label{fig:distance_headway}
\end{subfigure}\hfil

\begin{subfigure}{0.3\textwidth}
\includegraphics[width=\textwidth]{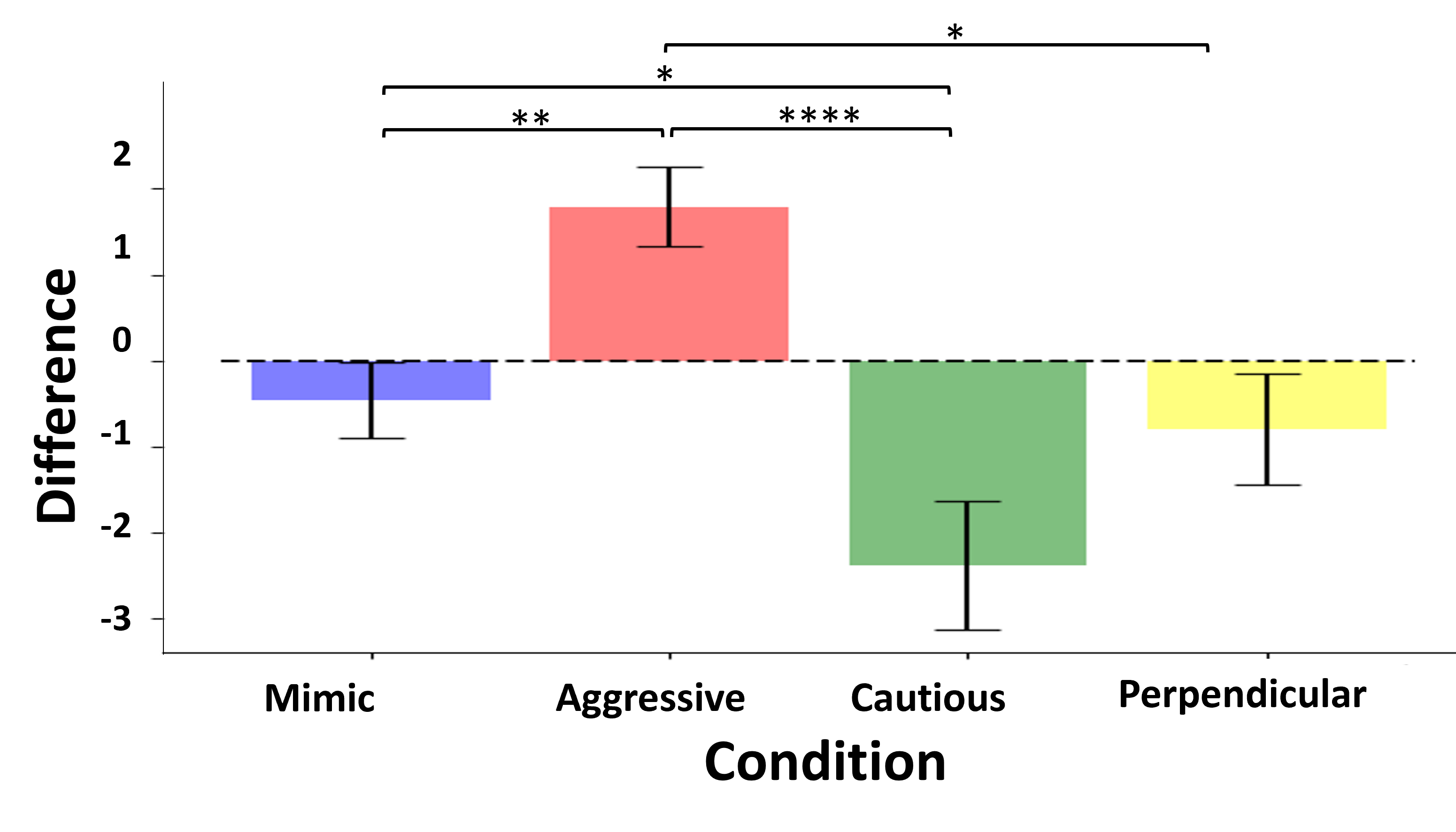}\vspace{-.2cm}
\caption{\footnotesize Mean number of lane changes}\label{fig:num_lane}
\end{subfigure}\hfil 
\begin{subfigure}{0.3\textwidth}
\includegraphics[width=\textwidth]{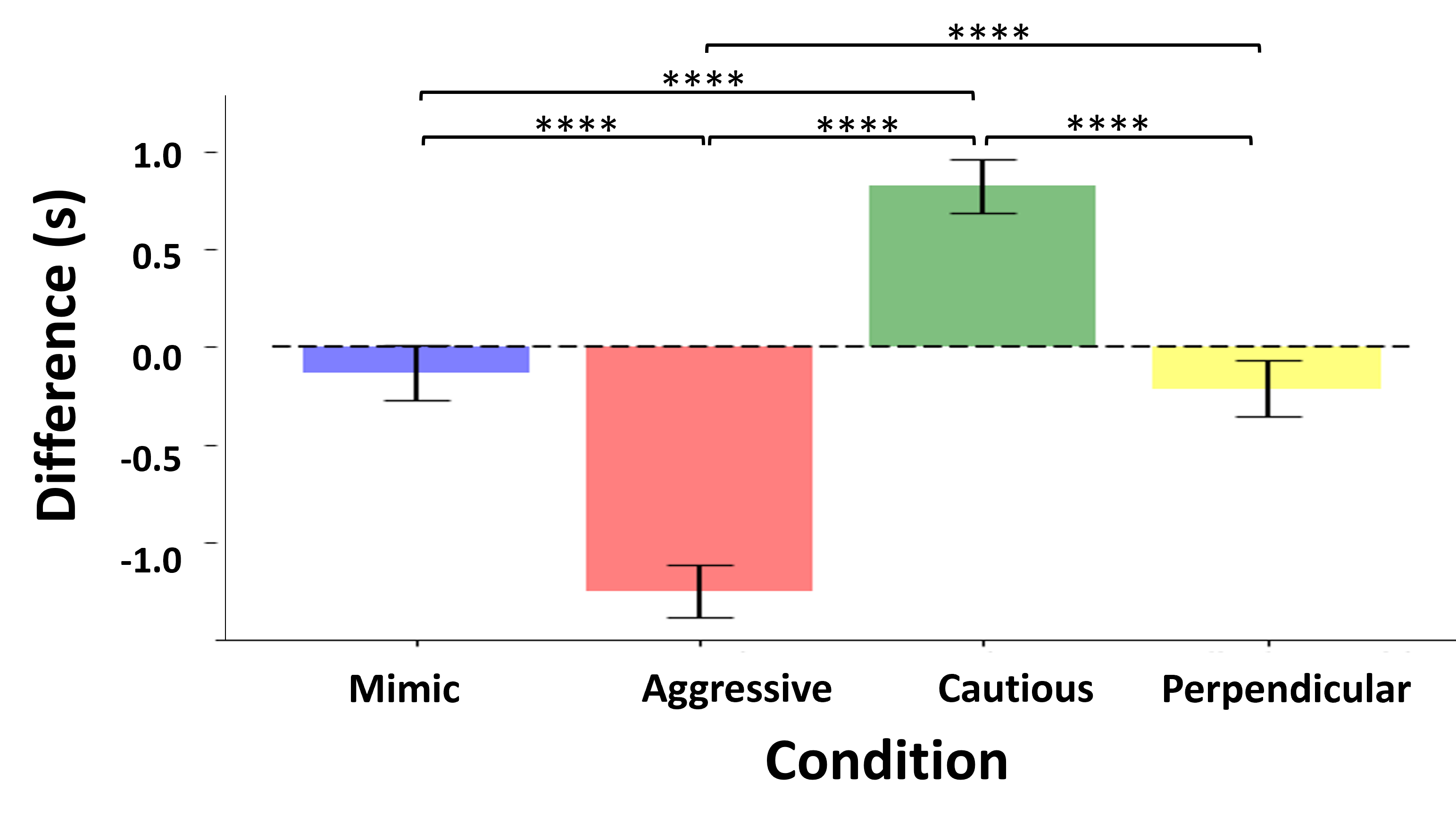}\vspace{-.2cm}
\caption{\footnotesize Time headway merge back}\label{fig:time_headway_back}
\end{subfigure}\hfil 
\begin{subfigure}{0.3\textwidth}
\includegraphics[width=\textwidth]{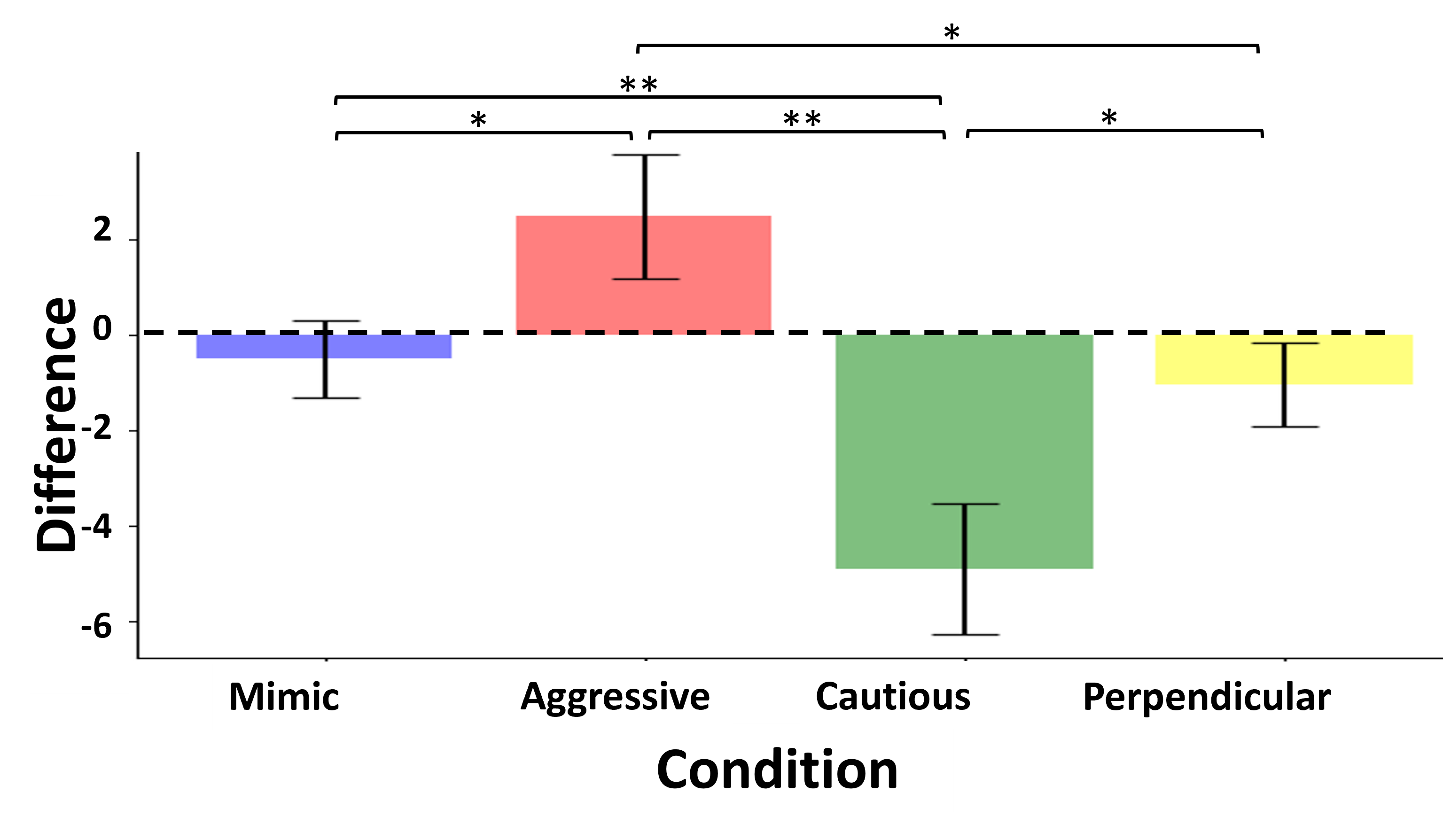}\vspace{-.2cm}
\caption{\footnotesize Subjective aggressive rating}\label{fig:subj_aggressive}
\end{subfigure}\hfil 

\vspace{-.1cm}
\caption[short caption]{This figure depicts the difference between the AV's driving style and the participants' driving style for our objective and subjective metrics. We show that both objectively and subjectively our approach can mimic an individual's driving style as well as modulate aggression.} \label{fig:results_objective}
\end{figure*}

\textbf{Study 2:} In the testing study, we freeze the network parameters, $\phi$, $\theta$, $\psi$, $\beta$, and $\alpha$ \emily{learned from Study 1} data. Participants fill out the pre-study surveys, complete a practice round, and then drive the vehicle in the highway domain. We collect their data to learn their embedding. This first part of the procedure \rsschange{mirrors the} procedure experienced by training participants. All participants next experience four AV conditions as described in Section \ref{sec:conditions}. After each condition, participants fill out surveys about their subjective perception of the AV (Section \ref{sec:metrics}).\looseness=-1


\subsection{Model Testing \revisiontwo{Study} Conditions}
\label{sec:conditions}
The  behaviors described below
are created by shifting a participant's embedding in the embedding space. Fig. \ref{fig:embeddings} shows the learned embedding space and how we choose the embedding to create the behavior for each of the conditions.  \revision{We hypothesize that Mimic will produce similar behavior relative to the participant's driving, Aggressive will produce more aggressive behavior and Cautious, less aggressive.}

\textbf{Mimic: } In \textit{Mimic}, we utilize the personalized embedding learned from the participant's data to produce driving behavior to mimic the participant's own driving style. 

\textbf{Aggressive: } In \textit{Aggressive}, we shift the participant's embedding  in the positive gradient of aggression \revision{(equivalent to 15 points on the ADB survey)} to produce more aggressive behavior while maintaining other characteristics of driving style (i.e., $S_\theta(\hat{w}^{(p)})=\hat{s}^{(p)}+15$). We constrain $\hat{s}^{(p)}+15$ to be no more than the largest possible score on the ADB survey (55 points).\looseness=-1 

\textbf{Cautious: } In \textit{Cautious}, we shift the embedding in the negative gradient of aggression ($S_\theta(\hat{w}^{(p)})=\hat{s}^{(p)}-15$) to produce less aggressive behavior while maintaining other characteristics of  style. We constrain  $\hat{s}^{(p)}-15$ to be no less than the smallest possible score on the ADB (11 points). 


\textbf{Perpendicular: } We include \textit{Perpendicular} to conduct an exploratory investigation into the behavior produced when we maintain the level of aggression but move the  embedding within the plane perpendicular to the aggressive gradient.
Our objective is to investigate which driving characteristics change as a result of this shift.
To select the embedding, we randomly sample a point along an ellipse on the plane one standard deviation away from the participant's embedding as shown by the gray points in Fig.
\ref{fig:embeddings}.
By doing so, we are able to keep the degree of aggression constant, while altering other aspects of driver style.
\revision{We hypothesize that Perpendicular will produce similarly aggressive behavior compared to the participant's driving but may modulate other factors not related to aggression.}


\subsection{Metrics}
\label{sec:metrics}
\emily{Participants in both Study 1 and Study 2 complete the pre-study surveys.} Only participants in \emily{Study 2} complete the post-trial surveys.  The surveys detailed below comply with the design guidelines outlined in  Schrum et al. \cite{Schrum2020FourStudies} and are validated from prior work when possible.

\textbf{Pre-study:}
The pre-study survey is intended to measure the participants' subjective attitudes towards AVs. We collect demographic information and Big-Five personality information via the Mini International Personality Item Pool \cite{cooper_confirmatory_2010}.  To measure a participant's aggressive driving style, we utilize the Aggressive Driving Behavior Scale \cite{AggressiveScale}.
We measure other aspects of driving style via the Multi-Dimensional Driving Style Inventory \cite{TAUBMANBENARI2016179} 
and measure experience with cars/racing games/AVs \cite{MINDMELD}, trust in AVs \cite{Jian1998}, perception of AVs \cite{Tennant2021}, and trust in automation \cite{TrustAutomation}.

\textbf{Post-trial: }
The post-trial surveys capture the participants' subjective attitudes towards each of the AV conditions. We measure perceived intelligence \cite{Bartneck2009}, competence \cite{ROSAS}, discomfort \cite{ROSAS}, and trust \cite{Jian1998}. We modify each of these subscales for AVs. Additionally, we create two custom scales to measure perceived similarity and aggressiveness relative to the participant's own driving style. 


\textbf{Objective Measures: }
In keeping with prior work \cite{Basu,Lee2004}, we measure various metrics to determine how similar the driving style of each condition is compared to the participant's own driving style. We investigate mean velocity and mean number of lane changes. We also measure mean headway time (the distance between the lead vehicle and ego divided by the speed of the ego when a lane change occurs), minimum headway distance (the minimum distance between the ego and lead vehicle before either the ego slows down or changes lanes),   distance headway merge back  (the distance  between the following vehicle and ego when merging back), and time headway merge back  (distance headway merge back divided by the speed when a lane change occurs).



\section{Results}
\label{sec:results}



\subsection{Analysis of Embedding Space and Aggressive Gradient} \rsschange{We first investigate if our embedding space is capable of representing and producing diverse driving styles and if the aggressive gradient correlates with relevant objective metrics. 
To investigate these questions, we project the learned embeddings of the test participants onto the line representing the gradient of aggression. We then analyze how driving style changes as a result of the position of the embedding along this line. We find that as we move along the aggressive gradient, the average velocity of the participant increases. The average velocity along the aggressive gradient ranges from 54.5 mph (in the most negative direction of the gradient) to 78.56 mph (in the most positive direction of the aggressive gradient. We find a strong correlation ($r=.49, p=.022$) between the embedding's position along the aggressive gradient and the average velocity of the participant. This finding suggests that, \rsschangetwo{in keeping with prior work \cite{driving_characteristics},} velocity is an important component of aggression within the embedding space. We find similar results for mean headway time ($r=-.46, p=.032$), distance headway merge back ($r=-.43, p=.046$), mean number of lane changes ($r=.47, p=.028$), and time headway merge back ($r=-.48, p=.025$). Lastly, we show that a participant's subjective aggressive rating of their own driving style strongly correlates with the position of their learned embedding along the aggressive gradient ($r=.92, p<.001$).} These findings \rsschangetwo{provide evidence} that our embedding space is capable of representing diverse driving styles and that aggressiveness objectively and subjectively increases as we move along the aggressive gradient.

\subsection{Algorithm Validation}
We next investigate \name{}'s ability to mimic end users' driving styles and produce more and less aggressive behavior in terms of both objective and subjective metrics. In our following analysis, we verify that data complies with assumptions before applying a parametric test. \rsschangetwo{We first investigate \name{}'s ability to accurately mimic driving style.  We find that the accuracy with which we are able to mimic the participant's velocity is 93.6\%, time headway is 80.2\%, distance headway merge back is 92.4\%, mean number of lane changes is 81.0\%, and time headway merge back  is  81.8\%. }

Fig. \ref{fig:results_objective} shows the differences in our objective and subjective metrics between the participant's driving style and the behavior produced by our four conditions. To determine if there are significant differences between  conditions for each of the  metrics, we conduct a repeated measures ANOVA  with Holm's post hoc  correction or a Friedman's test when the data fails  assumptions. We find that the difference between Mimic and the participant's driving is significantly less compared to Aggressive and Cautious for all objective metrics ($p<.001$)  (Fig  \ref{fig:average_vel} - \ref{fig:time_headway_back}).  \rsschange{We find that Aggressive maintains a higher velocity compared to Mimic. Additionally, as predicted by prior work \cite{Basu,Lee2004}, aggressive achieves a lower headway merge back time and headway merge back distance. Furthermore, Aggressive commits more lane changes compared to Mimic despite encountering the same number of leading vehicles. We find opposite results with the Cautious condition.}  \rsschangetwo{We  illustrate that the characteristics of our AV driving styles align with the characteristics indicative of aggression in prior work, suggesting that our approach can effectively modulate aggression with respect to one's own driving style \cite{Basu,Lee2004,driving_characteristics}.}

\andrewrss{I did not understand this figure until I got to section 6C which appears after it. I have added text to try to contextualize it better.}
\begin{figure}[h]
    \centering
    \includegraphics[width=.9\linewidth]{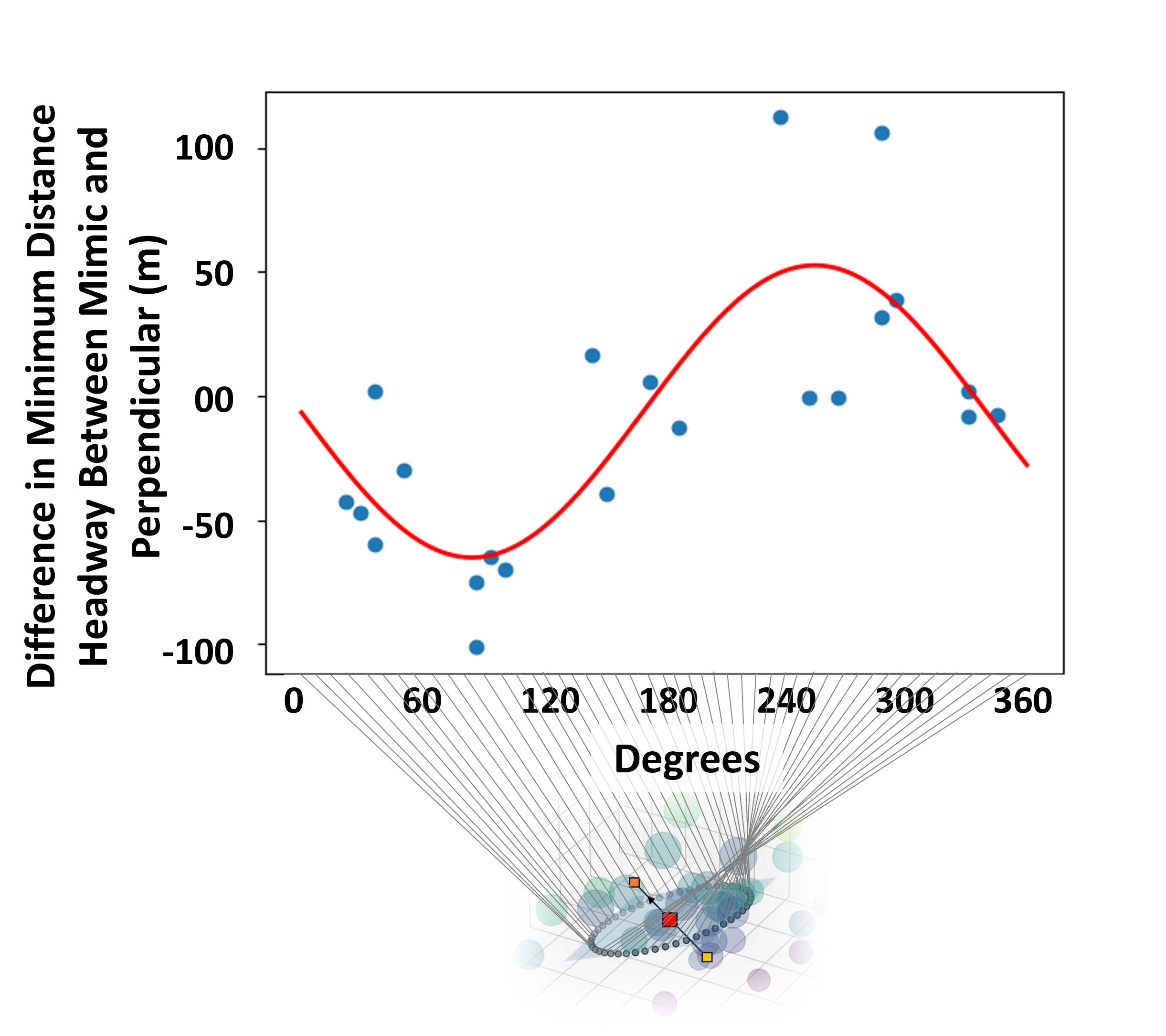}
    \caption{ This figure shows the changes in minimum headway distance as we move around the
ellipse within the plane perpendicular to aggression. Minimum headway distance was not significantly correlated with aggression (\ref{sec:maintainingOther}) and is modulated by moving in the plane perpendicular to aggression.}
    \label{fig:perp_dist_headway}
\end{figure}

Additionally, as shown in Fig. \ref{fig:subj_aggressive}, we find that participants rate Cautious as significantly less aggressive compared to Mimic ($p=.002$) and Aggressive as significantly more aggressive ($p=.017$). Furthermore, we find that Mimic and Perpendicular are rated as similarly aggressive compared to the participant's own driving.  
\revisiontwo{\textbf{Our objective and subjective results together \rsschangetwo{support our hypotheses} that 1) our approach is capable of mimicking driving style and 2), by shifting a participant's learned embedding along the aggressive dimension, we are able to produce objectively and subjectively more aggressive and cautious behavior.}}\looseness=-1

\subsection{Maintaining Other Aspects of Driving Style}
\label{sec:maintainingOther}
\rsschange{One of the goals of our approach is to modulate aggression while maintaining other aspects of driving style. If moving along the gradient of aggression modulates the aggressive aspect of the driving style, then we hypothesize that moving within the plane perpendicular to aggression will modulate other aspects of driving style unrelated to aggression. Interestingly, we found that minimum headway distance and \rsschangetwo{fraction of time in the left lane} were not significantly correlated with the embeddings position along the aggressive gradient. Moving along the gradient does not significantly alter minimum headway distance or fraction of time in the left lane, suggesting that, in our learned embedding space, these factors do  not play a large role in aggressiveness. Therefore, we predict that these aspects of driving will instead be modulated when we move perpendicular to the gradient of aggression. To test this hypothesis, in Fig \ref{fig:perp_dist_headway} we plot the difference in minimum headway distance between Mimic and Perpendicular versus the position around the ellipse that is depicted in Fig. \ref{fig:embeddings}. We find that minimum headway distance does in fact correlate with position around the ellipse \rsschangetwo{($r=.68, p<.001$).}} \rsschangetwo{We additionally find that the fraction of time in the left lane significantly correlates with position around the ellipse ($r=-.47, p=.025$).}

\andrewrss{Tried to find a bit of text here to poke reviewer 1 from ICRA who said "several features. However, can't these features all be explained with learning a mean desired velocity? Didn't figure it out, but if we want to poke this is the place to do it.}
We note that minimum headway distance is often associated with aggression \cite{XU202128}. However, this is most often the case when the ego vehicle is not capable of changing lanes and is instead forced to following a leading vehicle. We hypothesize in our work that minimum headway distance is not correlated with aggression because the participant can choose to change lanes at any point to pass a slower driver and therefore is not forced to maintain a following distance if they do not want to.



\begin{table}
\small
\begin{tabular}{ |c|c|c|c| } 
 \hline
 \textbf{Independent} & \textbf{Dependent}& \textbf{Statistic} & \textbf{p-value} \\ \hline
Conscientious & M-C Competence& $\rho(22)=-.71$ & $p<.001$ \\ \hline
 Conscientious  &M-C Intelligence & $\rho(22)=-.5$ & $p=.012$ \\ \hline
 Conscientious &M-C Discomfort & $r(22)=.46$ & $p=.024$  \\ \hline
 Conscientious  &M-C Trust & $\rho(22)=-.62$ & $p=.0011$ \\ \hline
 Conscientious & M-A Competence& $\rho(22)=-.51$ & $p=.011$ \\ \hline
 Conscientious  &M-A Intelligence & $\rho(22)=-.51$ & $p=.01$ \\ \hline
 Conscientious  &M-A Discomfort & $r(22)=.45$ & $p=.028$  \\ \hline
 Conscientious  &M-A Trust & $\rho(22)=-.48$ & $p=.0018$ \\ 
 \hline
  Openness  &M-A Discomfort & $\rho(22)=.49$ & $p=.015$ \\ \hline
   Similarity  &Trust & $\rho(94)=.16$ & $p=.001$ \\ \hline
    Similarity  &Intelligence & $\rho(94)=.34$ & $p<.001$ \\ 
 \hline
    Similarity  &Competence & $\rho(94)=.27$ & $p<.001$ \\ 
 \hline
 High-Velocity &M-A Intelligence & $\rho(94)=-.58$ & $p=.0031$ \\ 
 \hline
  High-Velocity &M-A Competence & $r(22)=-.44$ & $p=.03$ \\ 
 \hline
  High-Velocity &M-A Trust& $r(22)=-.43$ & $p=.036$ \\ 
 \hline
 
\end{tabular}\caption{This table shows our correlation analysis. M represents Mimic, A represents Aggressive, and C represents Cautious.}\label{table:correlation}
\end{table}
\subsection{\revisiontwo{Homophily}}
\begin{figure}[h]
    \centering
    \includegraphics[width=\linewidth]{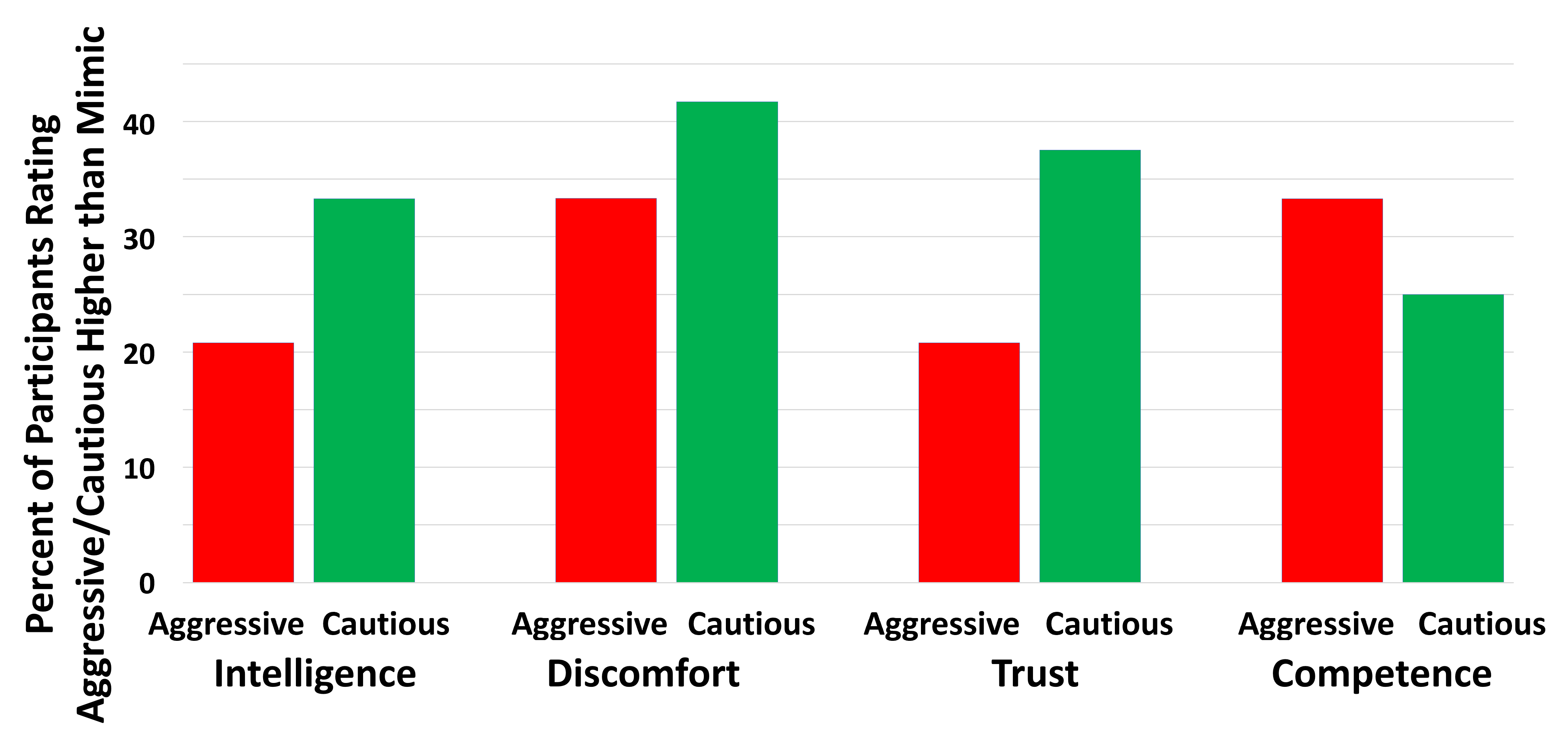}
    \caption{This figure shows the percent of participants who rated Aggressive and Cautious as better than Mimic in terms of each of our subjective metrics. }
    \label{fig:preference}
\end{figure}

\rsschange{As shown in Fig. \ref{fig:preference} not all participants preferred the Mimic condition. More than 20\% of participants preferred the Aggressive condition and more than 25\% of participants preferred the Cautious condition.  To explain this finding, we next}  explore the factors that modulate the effect of homophily (Table \ref{table:correlation}) to determine why some participants prefer a driving style different from their own. First we investigate if a participant's personality impacts their preference via a correlation analysis.  As shown in Table \ref{table:correlation}, we find a strong correlation between conscientiousness (i.e., the extent to which one is responsible and dependable \cite{Yu2021}) and the difference between a participant's perceived competence of Mimic compared to Cautious, suggesting that individuals higher in conscientiousness prefer a more cautious style to their own. This finding may explain why 62.5\% of participants rated Mimic as less than or equal in competence relative to Cautious. To further support the hypothesis that conscientiousness influences the effect of homophily, we  find that participants who are higher in conscientiousness rate a more cautious style as  significantly more intelligent, comfortable, and trustworthy compared to Mimic and significantly more competent, intelligent, comfortable, and  trustworthy   compared to Aggressive.  \looseness=-1

 We additionally  find that openness (the degree to which one is  broad-minded \cite{Yu2021}) correlates with the difference between a participant's comfort with Aggressive  compared to Mimic. This finding suggests that those who are more open to new experiences may  prefer a more aggressive AV and may explain why 37.5\% of participants rated Aggressive as causing greater comfort compared to Mimic.

 Prior work suggests that \textit{perceived similarity} to one's own driving style  is an important aspect of AV acceptance \cite{Basu}. To investigate this claim, we conduct a correlation analysis between perceived similarity and an end-users preference for the AV.  We find a positive correlation between perceived similarity and trust,   intelligence, and competence. 
This finding suggests that perceived similarity  should be taken into consideration when optimizing AV driving style.

Prior work demonstrated that one's own driving style may impact preference for an AV's style (e.g., more aggressive drivers prefer relatively less aggressive AVs) \cite{Basu,Yusof}. To investigate this question further, we conduct a correlation analysis between the dimensions of the Multi-Dimensional Driving Style Inventory \cite{TAUBMANBENARI2016179} and preference for Aggressive and Cautious compared to Mimic.   We find that participants who report a high-velocity driving style rate Aggressive to be significantly higher than Mimic in intelligence, competence,  and trustworthiness. This findings suggests that high-velocity drivers may prefer a more aggressive AV. Due to the contradictory findings with prior work, we aim to conduct a deeper analysis into how the specific dimensions of one's own aggressive style impact the effect of homophily in future work. 

We note that the results we present are an exploratory analysis and we do not claim to demonstrate a causal relationship between the subjective factors discussed above and homophily. However, our findings suggest that these factors warrant further investigation in future work. \revisiontwo{\textbf{Overall, our findings demonstrate that  personality traits, perceived similarity, and high-velocity driving style may be important factors in modulating the effect of homophily.}}\looseness=-1 





\section{Discussion}

\rsschange{Our results demonstrate the our \name{} framework is capable of both mimicking and modulating driving style by learning an embedding representing an end-user's own driving style. Given other relevant factors related to end-user characteristics, we can then tune this driving style to better match the preference of the end-user. Thus, while other approaches either directly mimic the end-user's own driving style or do not take into consideration the end-user's driving style at all, our approach is capable of integrating both information about an end-user's own driving style  and subjective characteristics that are predictive of the optimal AV driving style. }
\andrewrss{Is it worth noting that a three-vector is a pretty big dimension reduction compared to traditional control parameters? On the one hand a reviewer noted it, on the other, armed with a 3 vector, you'd have no idea how to use it without our network so I am on the fence.}

\rsschange{In our analysis, we show that our learned embedding space captures salient aspects of driving style and that the gradient of aggression correlates with objective and subjective aggressive metrics. An interesting aspect of our aggressive dimension is that this representation of aggression is not based on a pre-defined  or hand-crafted heuristic but is instead based upon end-users' perception of what is meant by aggressive driving. By defining aggression via this subjective metric, we are able to produce driving styles that are perceived to be more aggressive or more cautious by end-users.}

\rsschange{\rsschangethree{In our analysis of the effect of homophily,} we aim to determine the subjective factors that future work should consider when optimizing driving style. We show that simply mimicking an end-user's own driving style is often not preferred and that certain subjective characteristics may explain the discrepancy between an end-user's own driving style and their preferred AV driving style. By conducting a correlation analysis, we uncover several characteristics that impact homophily. We find that personality should be considered when determining the optimal driving style and that specifically,  conscientiousness and openness to experience are important factors. Additionally, participant's perception of their own driving style, e.g. self-reported high-velocity driving style may influence an individual's preference for a more aggressive driving style. We additionally find  that perceived similarity is a  relevant factor as supported in prior work \cite{Basu}. These findings provide us with insight into what factors should be considered when determining exactly how much and in which direction to shift an end-user's personalized embedding along the aggressive gradient so as to optimize driving style. 
}

\section{Limitations and Future Work}
\andrewrss{Added direct comparison language here}
In future work we aim to quantify the relationship between relevant subjective factors and the preferred level of aggression. By doing so, we will be able to determine exactly how much to shift an individual's embedding along the aggressive dimension so as to produce the optimal driving style for an individual. Additionally, we plan to investigate \name{}'s abilities to learn driving styles in  domains involving more traffic and the potential for more complex decision making. 

A limitation of our work is that we only recruited internal participants for Study 1. However, despite this limitation, our study comprises a more diverse population pool than many studies in human-robot interaction which typically recruit from a pool of college students \cite{hri_studies}. \revisiontwo{Another limitation is that the perceived similarity and aggressiveness surveys are not verified in prior work.} Additionally, because we only conduct a correlation analysis, we cannot conclude that the subjective factors are causally related to homophily. However, our results suggest that these factors are worthy of further investigation in future work.

\section{Conclusion}

We have presented \name{},  a novel framework
to personalize driving style and modulate aggressiveness. We demonstrated \name{}'s ability to reproduce an end-user's own driving style and investigated how the preference for one's own style is modulated by personality, self-reported driving style, and perceived similarity. To our knowledge, ours is the first framework to combine subjective metrics (i.e., the ADB survey) with end-user training data to produce a personalized AV controller. Our results indicate that personalizing AV control is a research area that merits further investigation and may provide a path towards greater AV acceptance.






\section*{ACKNOWLEDGMENT}





\bibliography{Bibtex.bib}
\bibliographystyle{plain}

\end{document}